\pdfoutput=1

\documentclass[11pt]{article}
\usepackage[utf8]{inputenc}

\usepackage[hyperref]{acl}
\usepackage{times}
\usepackage{latexsym}
\usepackage{amsmath}
\usepackage{booktabs}
\usepackage{graphicx}
\usepackage{multirow}
\usepackage{url}
\usepackage{tabularx}
\usepackage{enumitem}
\usepackage{caption}
\usepackage{float}

\usepackage{pifont} 
\usepackage{amssymb}

\usepackage{listings}
\usepackage{xcolor}
\usepackage{tcolorbox}

\usepackage{pgfplots}
\pgfplotsset{compat=1.18}

\lstdefinestyle{promptstyle}{
    basicstyle=\ttfamily\small, 
    breaklines=true,            
    breakatwhitespace=true,     
    frame=single,               
    framesep=5pt,
    rulecolor=\color{black!30}, 
    backgroundcolor=\color{gray!5}, 
    aboveskip=1em,
    belowskip=1em,
    columns=flexible,           
    showstringspaces=false      
}

\usepackage[T1]{fontenc}
\usepackage[utf8]{inputenc}
\usepackage{microtype}

\title{Infinite Problem Generator: Verifiably Scaling Physics Reasoning Data with Agentic Workflows}

\author{
    \textbf{Aditya Sharan}, \textbf{Sriram Hebbale}, \textbf{Dhruv Kumar} \\
    BITS Pilani, Pilani Campus, India \\
    \texttt{\{f20220674, f20220147, dhruv.kumar\}@pilani.bits-pilani.ac.in}
}

\begin{document}
\maketitle

\begin{abstract}
Training large language models for complex reasoning is bottlenecked by the scarcity of verifiable, high-quality data. In domains like physics, standard text augmentation often introduces hallucinations, while static benchmarks lack the reasoning traces required for fine-tuning. We introduce the Infinite Problem Generator (IPG), an agentic framework that synthesizes physics problems with guaranteed solvability through a "Formula-as-Code" paradigm. Unlike probabilistic text generation, IPG constructs solutions as executable Python programs, enforcing strict mathematical consistency. As a proof-of-concept, we release ClassicalMechanicsV1, a high-fidelity corpus of 1,335 classical mechanics problems expanded from 165 expert seeds. The corpus demonstrates high structural diversity, spanning 102 unique physical formulas with an average complexity of 3.05 formulas per problem. Furthermore, we identify a "Complexity Blueprint", demonstrating a strong linear correlation ($R^2 \approx 0.95$) between formula count and verification code length. This relationship establishes code complexity as a precise, proxy-free metric for problem difficulty, enabling controllable curriculum generation. We release the full \href{https://github.com/er-ads/ProblemGenerationAgent}{IPG pipeline}, the \href{https://huggingface.co/datasets/erads/ClassicalMechanicsV1}{ClassicalMechanicsV1} dataset, and our \href{https://er-ads.github.io/ProblemGenerationAgent/Physics_Evaluation_Report.html}{evaluation report} to support reproducible research in reasoning-intensive domains.
\end{abstract}

\begin{figure*}[ht]
    \centering
    \includegraphics[width=0.95\textwidth]{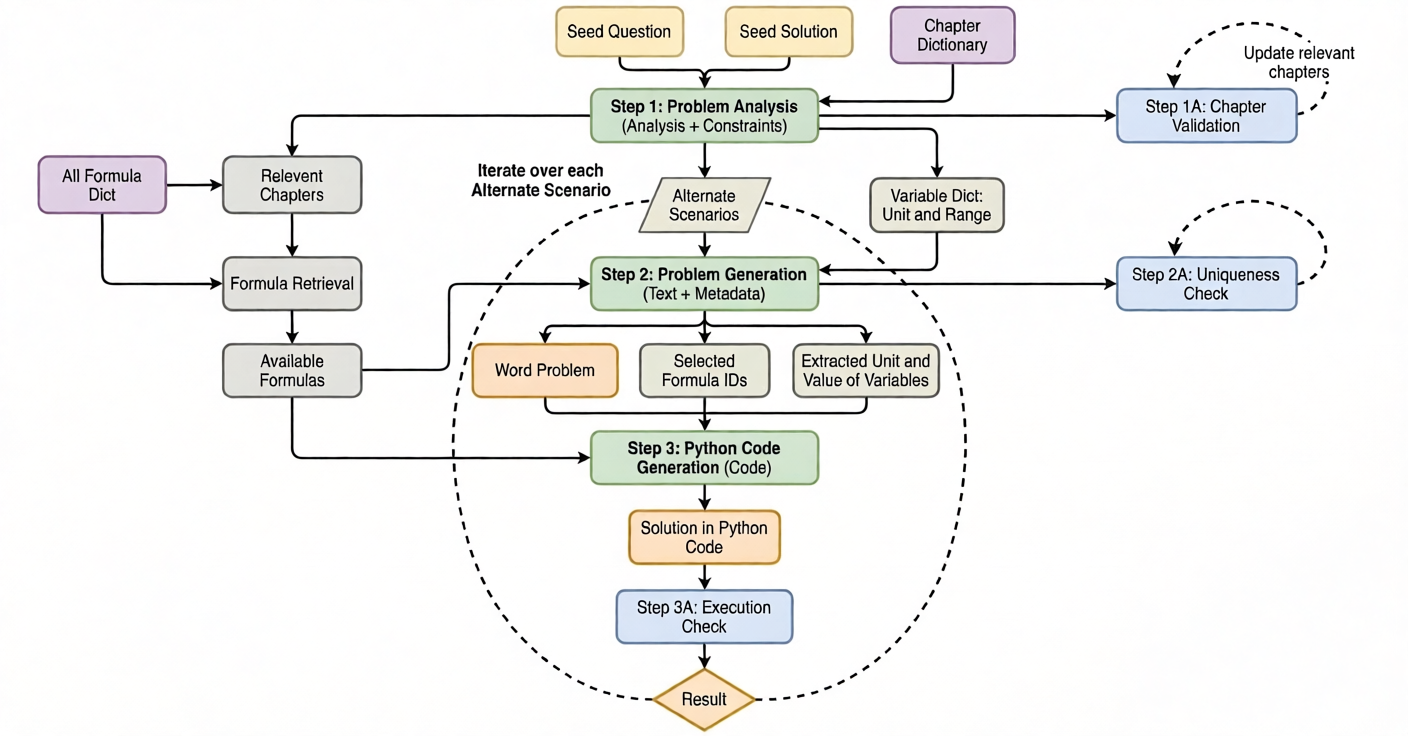}
    \caption{Overview of the proposed pipeline. \textbf{Problem Analysis} extracts constraints to guide \textbf{Constrained Generation}, while \textbf{Code-Based Verification} ensures the solvability of the resulting variations through Python execution.}
    \label{fig:pipeline_architecture}
\end{figure*}

\section{Introduction}

The adaptation of Large Language Models (LLMs) to specialized, high-reasoning domains remains fundamentally constrained by data scarcity. While general-purpose models excel at surface-level language tasks, domains requiring rigorous multi-step deduction--such as undergraduate physics and advanced mathematics--demand training data that web-scale corpora cannot adequately provide as noted by \citet{arora2023jeebench} and \citet{xu2025ugphysics}. Unlike natural language understanding tasks, physics problem solving requires identifying implicit constraints, selecting appropriate physical laws, and executing precise mathematical reasoning. Synthetic Dataset Generation (SDG) has emerged as a scalable solution \citep{ushio2022qgbench, long2024llm}, yet ensuring correctness in generated reasoning chains remains an open challenge.

We focus on physics problems at the level of the Joint Entrance Examination (JEE), a high-stakes entrance test attempted by over one million students annually in India. JEE problems are characterized by long-horizon, multi-step reasoning across tightly coupled concepts, fundamentally resisting shallow pattern-matching approaches \citep{arora2023jeebench}. We target this domain because it provides an ideal stress test for reasoning depth. Recent benchmarks such as \textbf{JEEBench} \citep{arora2023jeebench} and UGPhysics \citep{xu2025ugphysics} establish valuable evaluation standards; however, these datasets are static and designed primarily for testing. They lack the large-scale, diverse, fine-tuning-ready corpora with executable reasoning traces required to train robust reasoners, creating a persistent testing–training gap.

To address this gap, we introduce the \textbf{Infinite Problem Generator (IPG)}, an agentic framework for scalable and verifiable problem generation. Starting from expert-written seed problems, IPG systematically expands datasets by translating the underlying mathematical logic of a problem into multiple distinct physical contexts \citep{mathgenie_acl24, eqgen_ijcai24}. While surface narratives and numerical values vary, the core physical reasoning remains invariant, preserving educational value and logical rigor.

Crucially, we move beyond reliance on LLM self-consistency. As illustrated in \textbf{Figure 1}, IPG adopts a Formula-as-Code paradigm, treating physics equations not as text tokens but as executable Python functions. We employ a Program-of-Thought verification mechanism \citep{pal_2022, mirzadeh2024gsm}, requiring every generated problem to be solvable by an automatically generated Python script. This execution-based verification filters mathematically invalid generations and ensures that all problems in the resulting dataset admit correct and consistent solutions \citep{li2024planning}. Using this pipeline, we curate 165 high-quality seed problems from standard textbooks and expand them into a corpus of 1,335 verified problems, achieving approximately an $8\times$ expansion per seed.

Beyond generation, we analyze the structural determinants of problem difficulty. Our analysis reveals a reproducible \textit{Complexity Blueprint}: the number of integrated physics formulas correlates linearly ($R^2 \approx 0.95$) with the length and structural complexity of the corresponding solution code. This relationship provides a proxy-free mechanism for controlling difficulty, enabling curriculum-style dataset construction without human annotation.

Our contributions are threefold:

\begin{enumerate}
\item \textbf{Agentic Verification Framework (IPG):} We propose an agentic generation pipeline that couples narrative variation with code-execution verification, significantly mitigating mathematical hallucinations in synthetic physics data.
\item \textbf{ClassicalMechanicsV1 Dataset:} We release a training-ready corpus of 1,335 undergraduate-level physics problems with executable solution paths and verified numerical correctness.
\item \textbf{Complexity Blueprint:} We demonstrate a quantifiable relationship between structural problem properties and solution complexity, enabling difficulty-controlled problem generation for adaptive learning.
\end{enumerate}

\section{Related Work}

\begin{table*}[t]
\centering
\small
\begin{tabular}{lcccc}
\toprule
\mbox{\textbf{Method}} & \mbox{\textbf{Domain}} & \mbox{\textbf{Paradigm}} & \mbox{\textbf{Exec.\ Verify}} & \mbox{\textbf{Dataset}} \\
\midrule
\mbox{PAL \citep{pal_2022}} & \mbox{Math} & \mbox{Inference} & \mbox{Exec.\ inference} & \mbox{GSM-hard} \\
\mbox{MathGenie \citep{mathgenie_acl24}} & \mbox{Math} & \mbox{Augment} & \mbox{Exec.\ filtering} & \mbox{MathGenieLM} \\
\mbox{MetaMath \citep{yu2024metamath}} & \mbox{Math} & \mbox{Augment} & \mbox{Answer-match verify} & \mbox{MetaMathQA} \\
\midrule
\mbox{\textbf{IPG (Ours)}} & \mbox{\textbf{Physics}} & \mbox{\textbf{Agentic Seed}} & \mbox{\textbf{Gen.\ exec verify}} & \mbox{\textbf{ClassicalMechanicsV1}} \\
\bottomrule
\end{tabular}
\caption{Comparison of related works. IPG applies an agentic seed paradigm to physics with generative execution verification.}
\label{tab:method_comparison}
\end{table*}

\subsection{Synthetic Data and Question Generation}

Automatic Question Generation (AQG) has shifted from rigid template-based systems to flexible neural approaches \citep{eqgen_ijcai24}. While early systems like E-QGen required extensive domain-specific schemas, recent works leverage the generative priors of LLMs. Strategies such as back-translation \citep{mathgenie_acl24}, planning-first pipelines \citep{li2024planning}, and summarization-based filtering \citep{ushio2022qgbench} have improved fluency. However, most existing methods rely on \textbf{unstructured text corpora} or large knowledge bases as inputs. In domains like physics, where problems rely on precise initial conditions rather than general text, these methods struggle to maintain logical coherence. Our framework departs from text-scraping by operating on \textbf{expert-written seeds}, systematically expanding them via controlled logical variations rather than linguistic perturbation.

\subsection{Agentic Reasoning \& Verification }

Single-pass LLM generation is notoriously brittle for long-horizon reasoning \citep{long2024llm}. Concurrently, the \textbf{Program-of-Thought (PoT)} paradigm--exemplified by PAL \citep{pal_2022} and PoT \citep{pot_2022}--demonstrated that offloading logic to a Python interpreter significantly reduces calculation errors. Recent work has begun to merge these streams, using execution to filter synthetic data \citep{li2024planning, unlearning_arxiv25}. However, most prior work uses execution as a \textbf{post-hoc filter} (generating 100 samples and keeping the 10 that run). We integrate PoT directly into the \textbf{generation loop}, using execution traces to drive the expansion itself. This ensures that every generated variation is not just syntactically valid, but mathematically executable by design.

\subsection{Benchmarks vs. Training Resources }

The fragility of LLMs in physics is well-documented by benchmarks such as JEEBench \citep{arora2023jeebench}, UGPhysics \citep{xu2025ugphysics}, and PhysicsEval \citep{siddique2025physicseval}. Furthermore, \citet{mirzadeh2024gsm} showed that models often rely on surface-level pattern matching, failing when simple variables are permuted . Crucially, these benchmarks are \textbf{evaluative, not instructional}. They provide questions and final answers but lack the dense, step-by-step code traces required to fine-tune a model to \textit{reason}. Our work fills this void by providing a dataset that is "training-ready"--complete with intermediate code representations suitable for supervised fine-tuning and reinforcement learning approaches.

\subsection{Positioning }

Table 1 contextualizes our contribution. While approaches like MathGenie \citep{mathgenie_acl24} and MetaMath \citep{yu2024metamath} have successfully scaled mathematical data, they often result in variations that are structurally repetitive or lack physical context. Our work is the first to combine \textbf{seed-based expansion} with \textbf{executable verification} specifically for the physics domain, ensuring both structural diversity and rigorous correctness.

\section{Methodology}

We propose the \textbf{Infinite Problem Generator (IPG)}, an agentic synthetic data pipeline designed to facilitate domain adaptation in high-reasoning domains. IPG instantiates a multi-stage workflow that expands expert-written seed problems into verified training instances using executable reasoning.

As illustrated in Figure~\ref{fig:pipeline_architecture}, IPG follows a \emph{Generate--then--Verify} paradigm composed of three phases: \textbf{Problem Analysis}, \textbf{Constrained Generation}, and \textbf{Code-Based Verification}.

\subsection{Input Representation and Design Choices}

\subsubsection{Seed Tuple Definition}
IPG operates on a \textbf{Seed Tuple}
\[
S = \bigl(Q_{\text{seed}}, A_{\text{seed}}\bigr),
\]

where $Q_{\text{seed}}$ is an expert-authored physics question and $A_{\text{seed}}$ is its reference solution, sourced from standard undergraduate physics textbooks
\cite{hcv}.

\subsubsection{Executable Axioms (Formula-as-Code)}
Rather than representing equations as symbolic \LaTeX\ strings, IPG encodes physics formulas as \emph{executable axioms} implemented as Python functions. For example, the kinematic relation $v = u + at$ is represented as:
\begin{verbatim}
kinematics.final_velocity(u, a, t)
\end{verbatim}
This design choice is not intended as a constrained execution interface: IPG is restricted to invoking pre-defined, domain-validated axioms rather than generating free-form code. This enforces modularity, limits spurious operations, and enables runtime verification of numerical reasoning, extending prior Program-of-Thought approaches \citep{pal_2022,pot_2022} to a structured, domain-specific setting.

\subsection{Phase I: Problem Analysis and Context Expansion}

In Phase~I, IPG analyzes the seed tuple $S$ to construct the logical space required for controlled variation.

\paragraph{Underlying Principle Extraction:} IPG identifies the core physical principles governing $S$ and enumerates admissible real-world instantiations. For example, a seed involving angular acceleration of a pulley may be mapped to scenarios such as tire rotation, tape spools, fishing reels, or conveyor rollers. These mappings define narrative contexts without altering the underlying mechanics.

\paragraph{Concept Mapping via Chapter Dictionary:} As shown in Figure~\ref{fig:pipeline_architecture}, IPG queries a predefined \textbf{Chapter Dictionary} to map the extracted principles to relevant curriculum units. A seed originating in \emph{Rotational Motion} may activate additional chapters such as \emph{Circular Motion} or \emph{Rectilinear Motion}. The resulting union forms an \textbf{Available Formula Library} composed of executable axioms drawn from all activated chapters.  
For multi-step problems, IPG iteratively re-queries the Chapter Dictionary if the current library does not map to the seed solution, ensuring sufficient logical coverage.

\paragraph{Constraint Extraction:} IPG constructs a \textbf{Variable Dictionary} 
\[
V = \{(v_i, u_i, \mathcal{R}_i)\}_{i=1}^n,
\]
, where each variable $v_i$ is associated with a unit $u_i$ and a valid physical range $\mathcal{R}_i$ (e.g., $m > 0$, $\mu \in [0,1]$). These constraints guide parameter sampling and prevent physically implausible instantiations.

\subsection{Phase II: Constrained Problem Generation}

Given the expanded logical context, IPG generates $N$ variations (target $N=10$) while explicitly decoupling linguistic variation from numerical reasoning.

\paragraph{Narrative Round-Robin:} IPG cycles through the scenario set identified in Phase~I, generating a fixed number of problems per scenario. Each variation is required to be solvable using only the selected executable axioms, preventing hidden dependencies on unstated formulas.

\paragraph{Problem Signature and Uniqueness:} To detect and reject duplicates, each problem is assigned a \textbf{Problem Signature:}

\[
\Sigma = \bigl(\{ID_{f_1}, ID_{f_2}, \dots, ID_{f_k}\},\; v_{\text{target}}\bigr),
\]

where $ID_{f_j}$ denotes the identifiers of the invoked executable axioms and $v_{\text{target}}$ is the queried variable. Signatures are stored in a hash set; collisions trigger regeneration.

\paragraph{Difficulty Control:} Problem complexity is controlled by limiting the size of the active formula subset. IPG is asked to select between 3 and 5 axioms per instance, empirically encouraging multi-step reasoning and cross-chapter integration.

\subsection{Phase III: Solution Generation via Code Execution}

To significantly mitigate hallucination, IPG requires that each generated problem be accompanied by an executable Python solution. The solution is constructed by invoking only functions from the executable axiom library within a standardized \texttt{solve()} routine. Code is executed in a sandboxed environment and accepted only if it satisfies three criteria: (1) \textbf{Syntactic Validity}, ensuring the script executes without runtime errors; (2) \textbf{Numerical Solvability}, requiring that the output is finite (excluding $NaN$ or $\infty$); and (3) \textbf{Physical Sanity}, verifying that results satisfy basic sign and magnitude constraints (e.g., $t > 0$).

\subsection{Robustness and Efficiency}

IPG incorporates an internal retry loop informed by execution feedback. Failed attempts are re-prompted with structured error traces, enabling targeted correction. Across successful generations, IPG requires between 22 and 122 LLM calls per accepted problem, reflecting a deliberate trade-off favoring correctness and diversity over raw throughput.

\section{Experimental Setup}

\subsection{Dataset Construction}

\subsubsection{Seed Data Curation}
We curated 165 problem-solution pairs from \textit{Concepts of Physics} \cite{hcv}, specifically targeting Classical Mechanics (Chapters 3--10). This selection includes both textual exercises and ``Worked Out Examples'' to capture a wide range of pedagogical variations and difficulty levels.

\subsubsection{Formula Digitization}
The physics formula set was extracted from the \textit{Gyan Sutra} compilation. Utilizing \textbf{Gemini 2.5 Pro} \cite{gemini2.5}, we transformed these mathematical expressions into a structured Python library. This ``Formula-as-Code'' dictionary includes explicit docstrings and functional implementations to facilitate execution-based verification.

\subsection{Model Configuration and Baselines}

\textbf{Agentic Generation (IPG):} We utilized \textbf{Gemini 2.5 Flash} \cite{gemini2.5} to orchestrate all workflow phases. The agent operates within a fixed retry budget for automated error correction and signature collision handling. Notably, full formula definitions were embedded directly within the context window rather than retrieved via RAG, ensuring the model maintained full visibility into the implementation logic during synthesis.

\textbf{Zero-Shot Baseline:} To isolate the contribution of our agentic workflow, we generated a control dataset ($N=1,650$) using a single-prompt approach with the same base model. This baseline was designed to match the instructional density of the Agent output but lacked the intermediate analysis, constraint extraction, and iterative verification steps.

\subsection{Evaluation Framework}

\subsubsection{Intrinsic Dataset Metrics}
We evaluate the quality of the generated corpus using a number of intrinsic metrics including the following:
\begin{itemize}
    \item \textbf{Valid Execution Rate:} The percentage of problems where the generated code successfully produces finite, non-null numeric values.
    \item \textbf{Physical Sanity:} A filter detecting physically unrealistic values (e.g., mass $m < 0$ or astronomical displacements $|x| > 10^{15}$).
    \item \textbf{Signature Uniqueness:} The ratio of unique (Formula Set, Unknown Variable) tuples to the total population.
    \item \textbf{Lexical Diversity (TTR):} The Type-Token Ratio, calculated as $TTR = \frac{N_{\text{unique}}}{N_{\text{total}}} \times 100$, used to measure vocabulary richness.
\end{itemize}

\subsubsection{Extrinsic Stratified Audit}

We employed \textbf{Gemini 3} \cite{gemini3} as an independent judge for semantic validation. To probe the model's performance across varying reasoning depths, we utilized a stratified stress-test comprising three distinct tiers:

\begin{itemize}
    \item \textbf{Single-Step Baseline (0--1 Formulas):} Establishes a performance floor for conceptual knowledge retrieval.
    \item \textbf{Reliability Benchmark (2--3 Formulas):} Represents standard textbook-level complexity and serves as the control group.
    \item \textbf{Complexity Stress-Test (4--6 Formulas):} A long-tail subset designed to challenge context retention, variable tracking, and multi-step functional derivation.
\end{itemize}

The judge evaluated samples against a 12-point error taxonomy to isolate specific failure signatures, such as signature mismatches or physical impossibilities, quantifying the reliability trade-off at higher complexity tiers ($N \geq 4$).

\section{Results and Analysis}

We present a comprehensive analysis of the ClassicalMechanicsV1 corpus ($N=1,335$), quantifying structural diversity, reasoning depth, and code scalability. To ensure the dataset targets the ``reasoning gap,'' we initially generated 1,415 candidates and pruned 80 instances ($<6\%$) that required fewer than two deductive steps. Notably, our execution-based verification flagged only two problems in the final subset as numerically unstable (producing $\text{NaN}$ or $\infty$), confirming the robustness of the Generate-then-Verify paradigm.

\begin{table}[ht]
\centering
\small
\resizebox{\columnwidth}{!}{
\begin{tabular}{@{}lcccc@{}}
\toprule
\multirow{2}{*}{\textbf{Chapter}} & \multicolumn{2}{c}{\textbf{Seed Dataset}} & \multicolumn{2}{c}{\textbf{Generated}} \\ \cmidrule(lr){2-3} \cmidrule(lr){4-5}
 & \textbf{Count} & \textbf{\%} & \textbf{Count} & \textbf{\%} \\ \midrule
Kinematics & 30 & 18.18 & 185 & 13.86 \\
Newton's Laws & 16 & 9.70 & 149 & 11.16 \\
Friction & 11 & 6.67 & 87 & 6.52 \\
Work, Power, Energy & 21 & 12.73 & 200 & 14.98 \\
Circular Motion & 20 & 12.12 & 178 & 13.33 \\
Centre of Mass & 29 & 17.58 & 181 & 13.56 \\
Rigid Body Dynamics & 38 & 23.02 & 355 & 26.59 \\ \midrule
\textbf{Total} & \textbf{165} & \textbf{100.0} & \textbf{1,335} & \textbf{100.0} \\ \bottomrule
\end{tabular}
}
\caption{Comparative distribution of problems per chapter across seed and generated datasets.}
\label{tab:chapter_distribution_final}
\end{table}

\subsection{Structural Distribution \& Complexity}

We define ``Reasoning Complexity'' as the number of unique physics axioms (formulas) required to derive a solution. As shown in Table \ref{tab:formula_distribution}, the dataset exhibits a Gaussian-like distribution centered at a mode of 3 formulas (57.5\% of corpus). This clustering confirms the agent's proficiency at generating \textbf{Intermediate-Depth Reasoning}---chains that link multiple concepts (e.g., Kinematics $\rightarrow$ Energy) without becoming unwieldy.

\textbf{Foundational Instances (0--1 Formulas):} A small subset (5.6\%) serves as a conceptual baseline, testing definitions (e.g., Center of Mass coordinates) rather than derivations.

\textbf{Deep Reasoning (4--6 Formulas):} The ``Complexity Tail'' ($N=260$) represents long-horizon problems requiring the integration of up to 6 distinct physical laws, significantly exceeding the depth of standard benchmarks like GSM8K.

\begin{table}[ht]
\centering
\small
\begin{tabular}{@{}ccc@{}}
\toprule
\textbf{Formulas per Problem} & \textbf{Count} & \textbf{Percentage (\%)} \\ \midrule
0 & 38 & 2.69 \\
1 & 42 & 2.97 \\
2 & 261 & 18.44 \\
3 & 814 & 57.53 \\
4 & 198 & 13.99 \\
5 & 60 & 4.24 \\
6 & 2 & 0.14 \\ \midrule
\textbf{Total} & \textbf{1335} & \textbf{100.00} \\ \bottomrule
\end{tabular}
\caption{Distribution of generated physics problems by the number of unique formulas required for a solution.}
\label{tab:formula_distribution}
\end{table}

\subsection{Inter-Chapter Reasoning (Domain Mixing)}

A key indicator of quality is Domain Mixing---the ability to combine concepts from disparate chapters (e.g., Friction + Rotational Motion). While seed problems are chapter-specific, the IPG agent successfully breaks these boundaries. As detailed in Table \ref{tab:chapter_formulas}, the number of unique formulas used often exceeds the chapter's native library. For instance, Rigid Body Dynamics utilizes 53 unique formulas (vs. 20 native), indicating the agent actively pulls auxiliary laws from Kinematics and Energy to construct solvable scenarios. This confirms that the dataset contains integrated physics problems rather than isolated textbook drills.

\begin{table}[ht]
\centering
\small
\begin{tabular}{@{}lcc@{}}
\toprule
\textbf{Chapter} & \textbf{Library Total} & \textbf{Unique Used} \\ \midrule
Kinematics & 33 & 32 \\
Newton's Laws & 10 & 17 \\
Friction & 2 & 9 \\
Work, Power, Energy & 9 & 42 \\
Circular Motion & 20 & 25 \\
Centre of Mass & 18 & 46 \\
Rigid Body Dynamics & 20 & 53 \\ \bottomrule
\end{tabular}
\caption{Comparison of available library formulas versus unique formulas utilized per physics domain in the generated dataset. Higher usage indicates cross-domain mixing.}
\label{tab:chapter_formulas}
\end{table}

\subsection{The Complexity Blueprint: Code as a Proxy for Difficulty}

A central finding of this work is the ``Complexity Blueprint.'' We hypothesized that in a Program-of-Thought regime, code length is not random but a direct proxy for logical depth. Our analysis confirms a strong linear correlation ($R^2 \approx 0.953$) between the number of required formulas and the length of the verification code.

\begin{figure}[t]
\centering
\begin{tikzpicture}
\begin{axis}[
    width=\columnwidth,
    height=5.5cm,
    xlabel={\small Number of Formulas},
    ylabel={\small Avg. Code Length (Chars)},
    ymin=2000, ymax=4500,
    xmin=1.5, xmax=5.5,
    xtick={2,3,4,5},
    ytick={2000, 2500, 3000, 3500, 4000, 4500},
    legend pos=north west,
    xmajorgrids=true,
    ymajorgrids=true,
    grid style=dashed,
    footnotesize
]

\addplot[
    only marks,
    color=blue,
    mark=square*,
    mark size=2.5pt,
]
coordinates {
    (2, 2420)
    (3, 2635)
    (4, 3277)
    (5, 4011)
};

\addplot [
    domain=1.8:5.2,
    color=red,
    thick,
    samples=100,
]
{530*x + 1350}; 

\end{axis}
\end{tikzpicture}
\caption{The ``Complexity Blueprint.'' Code length scales linearly with formula count}
\label{fig:code_scalability}
\end{figure}
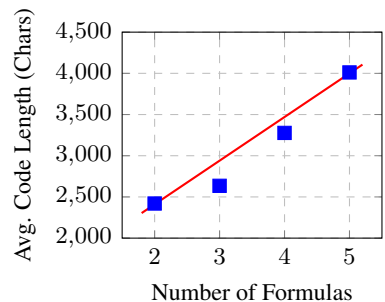

As visualized in Figure \ref{fig:code_scalability}, we observe a consistent ``cost'' of at least 250 characters per additional physical law. This linearity has two critical implications: first, \textbf{Hallucination is significantly mitigated}, as the model does not bloat code with irrelevant logic; length scales strictly with physical necessity. Second, \textbf{Controllable Generation}, where code length can serve as a reliable, proxy-free metric for estimating problem difficulty, enabling the generation of curriculum-style datasets without expensive human labeling.

\subsection{Lexical \& Semantic Diversity}

We calculated the Type-Token Ratio (TTR) to assess linguistic variety. The dataset achieves a TTR of 5.94. In the context of physics, this relatively focused vocabulary is a positive signal of domain adherence, reflecting the consistent use of precise technical terminology rather than generic synonyms. Table \ref{tab:unknown_variables} confirms this, showing high-frequency distribution across specific identifiers like \texttt{angular\_acceleration} and \texttt{normal\_force}, indicating correct domain mapping rather than simple pattern matching.

\begin{table}[ht]
\centering
\small
\begin{tabular}{lr}
\toprule
\textbf{Unknown Variable} & \textbf{Frequency} \\
\midrule
acceleration & 33 \\
displacement & 27 \\
mass & 23 \\
normal\_force & 22 \\
angular\_acceleration & 21 \\
work\_done & 21 \\
v & 20 \\
\bottomrule
\end{tabular}
\caption{Distribution of the top 7 target unknown variables.}
\label{tab:unknown_variables}
\end{table}

\subsection{Failure Mode Analysis}

We conducted a qualitative audit using a 12-point error taxonomy, revealing a distinct ``Fragility Shift'' as complexity increases (Table \ref{tab:failures_swapped}).

\textbf{The Reliability Zone (2--3 Formulas):} In this tier, validity exceeds 99\%. The primary ``error'' is the inclusion of \textit{Unused Variables} ($\approx 12\%$), such as providing atmospheric pressure in a basic gravity problem. We argue these are features, not bugs---they act as natural distractors that test a model's ability to filter irrelevant information.

\textbf{The Fragility Zone (4+ Formulas):} At high complexity, errors shift to \textit{Signature Mismatches} ($\approx 15\%$), where the agent correctly derives intermediate values but fails to chain them to the final target variable. This highlights specific limitations in current LLMs regarding maintaining long-horizon variable contexts, which this dataset is specifically designed to expose.

\begin{table*}[t]
\centering
\small
\begin{tabular}{cccc}
\toprule
\textbf{Dataset (Formulas)} & \textbf{Key Finding} & \textbf{Primary Category} & \textbf{Incidence} \\
\midrule
0--1 & Exhibit Triviality, not incorrectness & Math/Logical & $\sim$100\%  \\
2--3 & Unused distractor variables; correctly filtered & Variable Hallucination & $\sim$12\% \\
4--6 & Sound logic; random values may violate constraints & Logic/Text Alignment & $\sim$4--15\% \\
\bottomrule
\end{tabular}
\caption{Analysis of failure modes. The columns display the specific finding followed by the category categorization.}
\label{tab:failures_swapped}
\end{table*}

\subsection{Analysis of Low-Complexity and Pruned Instances}

Across the dataset, we observe 38 generated problems with zero formulas--all belonging exclusively to the \textit{Centre of Mass} chapter--and 42 single-formula problems distributed across multiple domains. Our analysis indicates that these cases are not inherently erroneous; rather, they correspond to single-step reasoning chains that fall below the complexity threshold of our multi-step benchmark.

\textbf{Domain-Specific Complexity Spikes:} A notable concentration of low-complexity problems in \textit{Centre of Mass} arises from the domain's core structure. Many problems reduce to direct coordinate or weighted-average calculations. While physically meaningful, these represent computationally shallow reasoning paths. Similarly, a substantial fraction of single-formula problems originates from \textit{Rigid Body Dynamics}, largely due to the prevalence of definitional relations such as $\tau = I\alpha$ and $L = I\omega$. When a scenario directly provides inertia and angular velocity, solving for angular momentum becomes a valid but trivial task, appearing as a ``defective'' instance within a multi-step reasoning context.

\textbf{Robustness of Dynamics and Energy:} In contrast, \textit{Newton's Laws} (Chapter 5) and \textit{Work, Power, and Energy} (Chapter 7) produced very few pruned instances (5 and 8 problems, respectively). These domains naturally encourage formula coupling, requiring multiple interacting quantities such as mass, force, and acceleration. Consequently, these chapters serve as more robust sources for the 2+ step reasoning chains necessary for effective domain adaptation.

\textbf{Quantitative Expansion Limits:} This behavior is reflected in Table~\ref{tab:chapter_distribution_final}. Although the agent targets a 10$\times$ expansion per chapter, this objective is not consistently achieved. For instance, while Chapter 5 reaches 9.31$\times$ and Chapter 7 reaches 9.52$\times$, \textit{Centre of Mass} only achieves 6.24$\times$ (181 problems from 29 seeds). This shortfall suggests that the agent frequently encounters the uniqueness constraint as a primary failure mode, particularly in domains where candidate problems fail to be meaningfully distinct. 

These observations suggest a potential link between the emergence of low-complexity instances and the model’s tendency to optimize against validation checks--prioritizing ``safe'' but simple problems to pass filters--rather than genuinely increasing reasoning depth.

\subsection{Deep Content Audit: Text-Code Alignment}

We assessed alignment between word problem narrative and Python logic to detect semantic hallucinations.

The audit revealed that narrative richness and reasoning complexity scale together. While the Defective cluster relied on sparse definitions, the Core cluster successfully integrated environmental details (e.g., atmospheric pressure) as effective distractors. For complex problems, the model correctly identified sophisticated principles, such as conservation of momentum being insufficient for pure rolling scenarios. However, strict physical constraints (e.g., friction requirements) remain the primary risk factor during execution.

\section{Limitations}

\paragraph{Semantic and Conservation Constraints:}
While Program-of-Thought verification ensures numerical correctness and unit consistency, it does not fully replace a symbolic physics engine. In highly complex scenarios ($N \geq 5$ interacting formulas), there remains a residual risk of semantic inconsistency, where a solution is mathematically valid but physically implausible (e.g., a car accelerating at $20g$ due to random parameter initialization). Although we enforce strict range constraints to mitigate such cases, future work could integrate formal constraint solvers (e.g., Z3) to enforce higher-order conservation laws more rigorously.

\paragraph{Visual Grounding:}
Our framework operates in the text–code modality. While the agent can describe physical setups (e.g., a block on an inclined plane), it does not generate corresponding visual diagrams. As reasoning models increasingly incorporate multimodal inputs, extending IPG with programmatic diagram synthesis (e.g., TikZ or SVG generation) represents an important direction for future work.

\paragraph{Domain and Axiomatic Scope:}
Our proof-of-concept focuses on Classical Mechanics. Extending the framework to domains such as Electromagnetism or Optics would require expanding the axiomatic library and supporting continuous field representations, which are less amenable to discrete formulation. Additionally, the Formula-as-Code paradigm captures algebraic reasoning effectively but does not yet model geometric intuition or first-principles calculus derivations.

\paragraph{Inference Cost:}
The generate-and-verify paradigm prioritizes correctness over efficiency. The iterative rejection loop--used to resolve execution errors and signature collisions--incurs higher computational cost per sample than lightweight text-based augmentation. Future work may incorporate lightweight solvability predictors or small-model filters to improve generation efficiency.

\section{Conclusion and Future Work}

We presented the \textbf{Infinite Problem Generator}, an agentic framework for addressing the scarcity of high-quality reasoning data in specialized domains. By decoupling narrative generation from numerical reasoning through \textbf{Program-of-Thought (PoT)} verification, we expanded 165 expert-written seed problems into 1,335 unique, executable variations, achieving a 99.85\% verification success rate. Our intrinsic analysis identified a reproducible \textit{Complexity Blueprint}, demonstrating a linear correlation between the number of integrated formulas and solution code length ($R^2 \approx 0.953$). This result suggests that code-based solution structure can serve as a proxy-free signal for problem difficulty, enabling controlled scaling of reasoning depth while reducing logical inconsistencies common in text-only generation.

Future work will focus on three directions:

\begin{enumerate}
    \item \textbf{Expanded Curriculum Coverage:} Extending beyond Classical Mechanics to other undergraduate physics domains such as Electromagnetism and Optics, and to adjacent sciences, by scaling the underlying formula library and adapting verification logic.
    
    \item \textbf{Multimodal Extensions:} Incorporating visualization modules capable of generating aligned diagrams (e.g., SVG or TikZ) alongside textual problem statements, supporting geometry- and diagram-intensive reasoning.
    
    \item \textbf{Adaptive Assessment:} Leveraging the proposed complexity signal to construct systems that dynamically assemble difficulty-controlled problem sets for curriculum design and adaptive testing.
\end{enumerate}

\section*{Acknowledgments}
The authors wish to acknowledge the use of ChatGPT, Claude and Gemini in improving the presentation and grammar
of the paper. The paper remains an accurate representation of the authors’ underlying contributions.

\bibliography{references}

\appendix

\section{Dataset Evaluation Metrics}
\label{sec:metrics}

We evaluated the generated dataset using the following metrics. For complete evaluation details and interactive visualizations, see our online report: \url{https://er-ads.github.io/ProblemGenerationAgent/Physics_Evaluation_Report.html}

\subsection{Structural Metrics}

\textbf{Total Problems:} Total number of physics problems in the dataset or chapter.

\textbf{Signature Uniqueness:} Percentage of distinct problem signatures computed as $\frac{\text{unique signatures}}{\text{total signatures}} \times 100$, where each signature represents a unique combination of formulas and unknown variable. Higher values indicate more diverse problem structures.

\textbf{Avg Formulas/Problem:} Mean number of physics formulas used per problem, computed as $\text{mean}(\text{formula counts per problem})$. Indicates problem complexity.

\textbf{Difficulty Level:} Categorical assessment based on average formulas per problem: Easy ($< 2$), Medium ($2{-}3$), or Hard ($> 3$).

\subsection{Diversity Metrics}

\textbf{Text Uniqueness:} Percentage of problems with unique wording computed as $\frac{\text{unique texts}}{\text{total texts}} \times 100$. Higher values indicate less repetition in problem statements.

\textbf{Duplicate Texts:} Number of problems with non-unique wording, calculated as $\text{total texts} - \text{unique texts}$. Lower is better for dataset diversity.

\textbf{Diversity (Type-Token Ratio / TTR):} Vocabulary richness measured as $\frac{\text{unique words}}{\text{total words}} \times 100$ across all problem texts. Higher values indicate more varied language and less repetitive wording.

\textbf{Unique Formulas:} Total count of distinct formula identifiers used across all problems, computed as $|\text{set}(\text{all formula IDs})|$. Indicates breadth of physics concepts covered.

\textbf{Unique Unknowns:} Total count of distinct unknown variables being solved for, computed as $|\text{set}(\text{all unknown vars})|$. Shows variety in problem objectives.

\textbf{Avg Word Count:} Mean number of words per problem statement, computed as $\text{mean}(\text{word counts})$. Indicates average problem description length.

\subsection{Quality Metrics}

\textbf{Valid Answers:} Percentage of problems with non-null numerical results, computed as $\frac{\text{problems with results}}{\text{total problems}} \times 100$. Should ideally be 100\%.

\textbf{Unrealistic Values:} Count of numerical results that are either extremely large $(|\text{value}| > 10^{15})$ or extremely small $(|\text{value}| < 10^{-15})$, which may indicate computational errors or physically implausible scenarios.

\textbf{Avg Code Length:} Mean character count of solution code snippets, computed as $\text{mean}(\text{len}(\text{code}))$. Provides insight into solution complexity.

\section{Failure Mode Taxonomy}
\label{sec:taxonomy}

We employed a 12-point taxonomy to systematically assess failure modes across all generated problems (formula counts: 0, 1, 2, 3, 4, 5, 6). Each problem was evaluated against the following categories:

\subsection{Structural Failures}

\textbf{1. Execution/Validation failures:} Generated code fails to execute or produces runtime errors.

\textbf{2. Missing required fields:} Problem JSON lacks essential fields such as problem text, formulas, or unknown variables.

\textbf{3. Formatting inconsistencies:} Non-standard formatting or structure that deviates from expected schema.

\subsection{Mathematical Failures}

\textbf{4. Insufficient formulas (0-1):} Problems requiring complex reasoning but using fewer than 2 formulas, resulting in trivial solutions.

\textbf{5. Syntax errors in code:} Python code contains syntax errors preventing execution.

\textbf{6. Null/unrealistic results:} Code executes but produces NaN, Inf, or physically implausible numerical values.

\textbf{7. Wrong formula IDs:} Formula identifiers do not match the Formula Dictionary or are misapplied to the problem context.

\subsection{Logical Failures}

\textbf{8. Signature mismatches:} Declared problem signature (formula set + unknown variable) does not align with actual solution logic.

\textbf{9. Variable issues:} Problems contain undefined variables, unused distractor variables, or variable naming inconsistencies.

\textbf{10. Physics impossibilities:} Solutions violate fundamental physics constraints (e.g., conservation laws, friction bounds $\mu > 1$).

\subsection{Semantic Failures}

\textbf{11. Hallucinations:} Problem narrative contains fabricated physical scenarios or introduces concepts not grounded in the formula set.

\textbf{12. Minor template artifacts:} Residual template text or placeholder values from generation prompts.

\textbf{13. Low uniqueness:} Near-duplicate problems with minimal variation from existing problems in the dataset.

\section{Agent Prompt Templates}

Below are the system prompts used for the Problem Generator agent.

\subsection{Step 1: Problem Analysis}
\label{sec:prompt_analysis}

\begin{lstlisting}[style=promptstyle, basicstyle=\ttfamily\scriptsize]
sys_call1 = """
You are a physics problem analyzer. Your task is to analyze a physics question and its solution, then extract key information.

INPUT:
- Chapter Dictionary: {chapters_json}
- Physics Question: {question}
- Solution: {solution}

TASK:
Analyze the question and solution, then provide:

1. RELEVANT CHAPTERS: Identify exactly 2 chapters from the provided description in Chapter Dictionary that are most relevant to solving this problem.

2. VARIABLES: List all physical quantities (variables) involved in the problem. For each variable, specify:
   - A reasonable range of values [minimum, maximum]
   - The SI unit of measurement

3. ALTERNATE SCENARIOS:
   - Generate 6 real-world scenarios that could be used to create physics problems using the same core concepts as the original question.
   - Only the physical setting should change, while the underlying ideas remain the same.
   - Each scenario should be 1-2 sentences and should only change the physical situation.

OUTPUT FORMAT (JSON):
{
  "relevant_chapters": ["chapter_name_1", "chapter_name_2"],
  "variables": {
    "variable_name": {
      "range": [min_value, max_value],
      "unit": "unit_string"
    }
  },
  "alternate_scenarios": [
    "scenario description 1",
    "scenario description 2",
    "scenario description 3",
    ...
    "scenario description 6"
  ]
}

Provide only the Strictly JSON output, no additional explanation, not any other characters preceding or following the JSON.
"""
\end{lstlisting}

\subsection{Step 1A: Formula Verifications}
\label{sec:prompt_analysis}

\begin{lstlisting}[style=promptstyle, basicstyle=\ttfamily\scriptsize]
sys_call1a = '''
You are a physics formula verifier. Your task is to check if a given set of formulas is sufficient to solve a physics problem.

INPUT:
- Original Solution: {solution}
- Identified Chapters: {identified_chapters}
- All Formulas Chapterwise: {all_chapters_json}

TASK:
"Check if the solution can be fully solved using only the formulas available in the chapters listed in Identified Chapters."
1. For each step in the original solution, attempt to map it carefully and thoroughly to one or more formulas from the chapters whose names appear in Identified Chapters.
- Ensure you do not incorrectly return ``NO'' if a valid mapping actually exists.
2. If all steps can be matched with these formulas, output YES.
3. If any step cannot be mapped, output NO and identify a missing chapter from the complete chapter list.
- The missing chapter must be distinct from those already present in Identified Chapters.
- Choose the most relevant chapter that contains the formula or concept needed for the unmapped step.

OUTPUT FORMAT (JSON):
If formulas are sufficient:
{{
  "status": "YES"
}}

If formulas are NOT sufficient:
{{
  "status": "NO",
  "missing_chapter": "chapter_name",
  "reason": "2-line explanation of what formula/concept is missing"
}}

Provide only the Strictly JSON output, no additional explanation, not any other characters preceding or following the JSON.
'''
\end{lstlisting}

\subsection{Step 2: Constrained Problem Generation}
\label{sec:prompt_analysis}

\begin{lstlisting}[style=promptstyle, basicstyle=\ttfamily\scriptsize]
sys_call2 = '''
You are a physics problem generator. Your task is to create a new physics word problem based on provided scenarios and formulas.

INPUT:
- Available Formulas: {available_formulas}
- Alternate Scenario: {alternate_scenarios}
- Variables and Ranges: {variables}
- Previous Problems (avoid duplicates): {previous_problems}

TASK:
Generate a NEW physics word problem following these rules:

1.Pick the alternate scenario.
2.Select 3-5 formulas from the available formulas (use their formula_ids).
- The most important requirement is that the physical situation described in the word problem maps correctly to the selected formulas. 
There must be no conceptual mismatch between the scenario and the equations used.
- The chosen formulas do not all need to come from the same chapter -- you may select any formulas from the available_formulas list.
3.Create a word problem fully based on the chosen formulas and scenario.
4.The problem must be solvable using only the selected formulas -- no additional equations should be needed.
5.Assign specific numerical values to all variables:
- Each value must fall within its allowed range.
- Mark exactly one variable as "NaN" (the unknown to be solved).
6. Ensure the new problem is meaningfully different from previous ones.

OUTPUT FORMAT (JSON):
{{
  "word_problem": "Complete problem statement as text",
  "formula_ids": ["formula_id_1", "formula_id_2"],
  "variables": {{
    "variable_name_1": {{
      "value": numerical_value,
      "unit": "unit_string"
    }},
    "unknown_variable": {{
      "value": "NaN",
      "unit": "unit_string"
    }}
  }}
}}

IMPORTANT:
- The word problem should be a complete, standalone problem that a student could solve
- Include all necessary context and information in the problem statement
- Use clear, simple language
- Exactly ONE variable must have value "NaN"

Provide only the Strictly JSON output, no additional explanation, not any other characters preceding or following the JSON.
'''
\end{lstlisting}

\subsection{Step 2A: Problem Analysis}
\label{sec:prompt_analysis}

\begin{lstlisting}[style=promptstyle, basicstyle=\ttfamily\scriptsize]
sys_call2a = '''
You are a physics problem generator. Your previous attempt had an issue. Generate a corrected physics word problem.

PREVIOUS ERROR: {error_message}

INPUT:
- Available Formulas: {available_formulas}
- Alternate Scenario: {alternate_scenarios}
- Variables and Ranges: {variables}
- Previous Problems (avoid duplicates): {previous_problems}

TASK:
Generate a new physics word problem that corrects the previous mistake.

Guidelines:
1.Pick the scenario.
2.Select 3-5 formulas from the available formulas (use their formula_ids).
- The selected formulas must logically match the chosen scenario 
- there should be no mismatch between the physical situation and the equations used.
3.Create a word problem fully based on the chosen formulas and scenario.
- Difficulty level: The problem should be at least JEE Mains level, requiring clear conceptual understanding and 1-3 steps of reasoning.
4.The problem must be solvable using only the selected formulas -- no additional equations should be needed.
- Optionally state direction/sign convention (downward positive) to avoid sign ambiguity.
5.Assign specific numerical values to all variables:
- Each value must fall within its allowed range.
- Mark exactly one variable as "NaN" (the unknown to be solved).
6. Ensure the new problem is meaningfully different from previous ones.
7. Explicitly fix the error from the last version.

Clarity Requirement:
Avoid any ambiguity regarding: Which variable is being asked for, and Which variable corresponds to each given numerical value.
However, do not spoon-feed the answer -- the problem may still include a small element of inference or interpretation, as in real exam-style questions.

OUTPUT FORMAT (JSON):
{{
  "word_problem": "Complete problem statement as text",
  "formula_ids": ["formula_id_1", "formula_id_2"],
  "variables": {{
    "variable_name_1": {{
      "value": numerical_value,
      "unit": "unit_string"
    }},
    "unknown_variable": {{
      "value": "NaN",
      "unit": "unit_string"
    }}
  }}
}}

Provide only the Strictly JSON output, no additional explanation, not any other characters preceding or following the JSON.
'''
\end{lstlisting}

\subsection{Step 3: Solution Code Generation}
\label{sec:prompt_analysis}

\begin{lstlisting}[style=promptstyle, basicstyle=\ttfamily\scriptsize]
sys_call3 = '''
You are a Python code generator for physics problems. Your task is to write code that solves a physics word problem.

INPUT:
- Word Problem: {word_problem}
- IDs for Allowed Formulas: {formula_ids}
- Variables: {variables_dict}
- All Available Formulas: {available_formulas}

TASK:
Write Python code that solves for the unknown variable in the problem.  

REQUIREMENTS:
1. Import only: math, numpy (if needed)
2. Define all known variables from the variables dictionary
3. Use ONLY the formulas whose formula_ids are specified in the input
4. For each of mentioned Formula IDs, while accessing, directly copy their corresponding "python_code" from available_formulas
5. Use these copied functions by calling them inside the solve() function
6. Solve for the unknown variable (the one with value "NaN")
7. Return a single float value as the answer
8. Include try-except error handling
9. Define everything inside a function called solve()

CODE STRUCTURE:
```
import math
# import numpy as np  # only if needed

# As-it-is Copied functions from available_formulas based on the given formula_ids

def solve():
    try:
        # Define known variables
        variable_1 = value_1
        variable_2 = value_2

        # Use the provided formula functions
        # result = formula_function(...)

        # Return the computed answer
        return answer
    except Exception as e:
        return None
```

OUTPUT:
Provide ONLY the complete Python code. No explanations, no markdown formatting, just the raw code.
'''
\end{lstlisting}

\subsection{Step 3A: Solution Code Correction}
\label{sec:prompt_analysis}

\begin{lstlisting}[style=promptstyle, basicstyle=\ttfamily\scriptsize]
sys_call3a = '''
You are a Python code generator for physics problems. Your previous code failed. Generate corrected code.

PREVIOUS ERROR: {error_message}

INPUT:
- Word Problem: {word_problem}
- IDs for Allowed Formulas: {formula_ids}
- Variables: {variables_dict}
- All Available Formulas: {available_formulas}

TASK:
Write Python code that solves for the unknown variable in the problem.

REQUIREMENTS:
1. Import only: math, numpy (if needed)
2. Define all known variables from the variables dictionary
3. Use ONLY the formulas whose formula_ids are specified in the input
4. For each of mentioned Formula IDs, while accessing, directly copy their corresponding "python_code" from available_formulas
5. Use these copied functions by calling them inside the solve() function
6. Solve for the unknown variable (the one with value "NaN")
7. Return a single float value as the answer
8. Include try-except error handling
9. Define everything inside a function called solve()
10. FIX THE PREVIOUS ERROR: {error_message}

CODE STRUCTURE:
```
import math
# import numpy as np  # only if needed

# As-it-is Copied functions from available_formulas based on the given formula_ids

def solve():
    try:
        # Define known variables
        variable_1 = value_1
        variable_2 = value_2

        # Use the provided formula functions
        # result = formula_function(...)

        # Return the computed answer
        return answer
    except Exception as e:
        return None
```

OUTPUT:
Provide ONLY the complete Python code. No explanations, no markdown formatting, just the raw code.
'''
\end{lstlisting}

\section{Dataset Samples}

Below are the raw data from some samples of the generated dataset.

\subsection{0-Formula Problem from Centre of Mass}

\begin{lstlisting}[style=promptstyle, basicstyle=\ttfamily\scriptsize]
{
  "chapter": "9.Centre of Mass",
  "word_problem": "A miniature freight train car with a mass of 50.0 kg is coasting along a track at an initial velocity of 10.0 m/s. It collides head-on with a stationary miniature caboose...",
  "execution_result": "0.5",
  "signature": "fids=[]|unknown=coefficient_of_restitution_e",
  "formula_ids": [],
  "unknown_var": "coefficient_of_restitution_e",
  "variables": {
    "mass_m": { "value": 50.0, "unit": "kg" },
    "initial_velocity_v": { "value": 10.0, "unit": "m/s" },
    "final_velocity_u1": { "value": 2.5, "unit": "m/s" },
    "final_velocity_u2": { "value": 7.5, "unit": "m/s" },
    "coefficient_of_restitution_e": {
      "value": null, "unit": "dimensionless"
    }
  },
  "code": "import math

    def solve():
        try:
            # Define known variables
            mass_m1 = 50.0  # mass of freight train
            mass_m2 = 50.0  # mass of caboose
            v1_init = 10.0  # init vel of freight train
            v2_init = 0.0   # init vel of caboose
            v1_final = 2.5  # final vel of freight train
            v2_final = 7.5  # final vel of caboose

            # Formula for coefficient of restitution (e):
            # e = (relative vel after) / (relative vel before)
            # e = (u2 - u1) / (v1 - v2)

            num = v2_final - v1_final
            den = v1_init - v2_init

            coeff_e = num / den

            return coeff_e
        except Exception as e:
            return None",
  "validation_result": {
    "valid": true,
    "unknown_var": "coefficient_of_restitution_e"
  }
}
\end{lstlisting}

\subsection{6-Formula Problem from Rigid Body Dynamics}

\begin{lstlisting}[style=promptstyle, basicstyle=\ttfamily\scriptsize]
{
  "chapter": "10.Rigid Body Dynamics",
  "word_problem": "A dart with a mass of 0.015 kg is thrown horizontally...",
  "execution_result": "0.12492187503051848",
  "signature": "fids=[10_A...]|unknown=h",
  "formula_ids": ["10_I", "10_A", "10_Q", "10_R", "9_K", "8_C"],
  "unknown_var": "h",
  "variables": {
    "m": { "value": 0.015, "unit": "kg" },
    "M": { "value": 12.0, "unit": "kg" },
    "R": { "value": 0.25, "unit": "m" },
    "v0": { "value": 40.0, "unit": "m/s" },
    "h": { "value": null, "unit": "m" }
  },
  "code": "import math

    def calculate_inertia_solid_cylinder_axis(
        mass: float, radius: float) -> float:
        return (1 / 2) * mass * radius**2

    def calculate_moment_of_inertia_discrete(
        masses: list[float], radii: list[float]) -> float:
        if len(masses) != len(radii):
            raise ValueError('Masses/radii length mismatch')
        return sum(m * r**2 for m, r in zip(masses, radii))

    def calculate_com_velocity_1d(
        masses: list[float], velocities: list[float]) -> float:
        total_mass = sum(masses)
        if total_mass == 0: return 0.0
        tot_mom = sum(m*v for m,v in zip(masses, velocities))
        return tot_mom / total_mass

    def solve():
        try:
            # Define known variables
            m = 0.015   # Dart mass in kg
            M = 12.0    # Log mass in kg
            R = 0.25    # Log radius in meters
            v0 = 40.0   # Initial speed in m/s

            # Step 1: Calculate Velocity of COM
            V_cm_system = calculate_com_velocity_1d(
                masses=[m, M], velocities=[v0, 0.0])

            # Step 2: Quadratic coefficients from conservation
            # m * h^2 - (m + M) * R * h + (1/2) * M * R^2 = 0
            a_coeff = m
            b_coeff = - (m + M) * R
            c_coeff = (1/2) * M * R**2

            # Solve quadratic for h
            disc = b_coeff**2 - 4 * a_coeff * c_coeff
            if disc < 0: return None

            h_sol1 = (-b_coeff + math.sqrt(disc)) / (2*a_coeff)
            h_sol2 = (-b_coeff - math.sqrt(disc)) / (2*a_coeff)

            # Check physical validity (within radius R)
            if 0 <= h_sol2 <= R + 1e-9:
                return h_sol2
            elif 0 <= h_sol1 <= R + 1e-9:
                return h_sol1
            else:
                return h_sol2 

        except Exception as e:
            return None",
  "validation_result": {
    "valid": true,
    "unknown_var": "h"
  }
}
\end{lstlisting}

\section{Downstream Evaluation of ClassicalMechanicsV1}
\label{app:downstream}

To assess the real-world utility of \textsc{ClassicalMechanicsV1} and
to address the terminology ambiguity in the original submission, we
clarify that while the dataset is described as ``fine-tuning-ready,''
our evaluation demonstrates its primary utility as a \textbf{rigorous
benchmark} for assessing physics reasoning capabilities.

\subsection{Experimental Protocol}

We evaluated Qwen3-14B in a zero-shot setting on a deterministically
sampled (seeded) subset of \textsc{ClassicalMechanicsV1} and on the
physics subset of JEEBench~\cite{arora2023jeebench}.  All evaluation
subsets are fixed at up to 123~problems; if a dataset contains fewer
than 123 items, all available items are used.

\paragraph{Dataset construction.}
\begin{itemize}
  \item \textbf{JEEBench (Physics):} physics subset only
        \cite{arora2023jeebench}.
  \item \textbf{ClassicalMechanicsV1:} 123 problems sampled
        deterministically from our dataset.
\end{itemize}

\paragraph{Results.}

\begin{table}[h]
  \centering
  \begin{tabular}{lrrr}
    \toprule
    \textbf{Dataset} & \textbf{Correct} & \textbf{Total} &
    \textbf{Accuracy} \\
    \midrule
    JEEBench (Physics)   & 59 & 123 & 47.97\% \\
    ClassicalMechanicsV1 & 43 & 123 & 34.96\% \\
    \bottomrule
  \end{tabular}
  \caption{Zero-shot performance of Qwen3-14B on
  \textsc{ClassicalMechanicsV1} versus the JEEBench physics subset.}
  \label{tab:downstream}
\end{table}

\subsection{Interpretation}

These results serve four purposes.  First, the fact that a capable
model achieves non-trivial accuracy on \textsc{ClassicalMechanicsV1}
provides an indirect signal of physical plausibility---problems that
were physically incoherent or unsolvable would not admit meaningful
performance.  Second, the results demonstrate real-world applicability:
\textsc{ClassicalMechanicsV1} functions as a meaningful evaluation
surface alongside established benchmarks.  Third, they demonstrate the
\textbf{feasibility of generating reasoning traces from the dataset}
using a capable model.  Fourth, and most importantly, because we provide
executable Python solutions for every single question, these generated
reasoning traces can be rigorously verified, ensuring the model is not
relying on hallucinated steps to arrive at a correct final answer.  This
represents a significant methodological improvement in generating
verifiable Chain-of-Thought data for SFT and RL pipelines.

\paragraph{Breakdown of JEEBench performance.}

A granular breakdown of performance on JEEBench highlights the need for
datasets like ours.  The model's accuracy was heavily inflated by
constrained answer spaces (scoring 68.2\% on Integer questions
vs.\ 48.5\% on open-ended Numerics) and by process-of-elimination
(scoring 55.6\% on Single-MCQ questions vs.\ 31.7\% on
Multiple-MCQ questions).  Because \textsc{ClassicalMechanicsV1} relies
on executable Python code generation rather than multiple-choice
selection, it strips away these testing artifacts, offering a more
robust and un-gameable metric of true physical reasoning.

Furthermore, JEEBench is widely recognised as an exceptionally
challenging benchmark composed of IIT JEE-Advanced problems, where
``long-horizon reasoning on top of deep in-domain knowledge is
essential'' \cite{arora2023jeebench}.  The fact that Qwen3-14B
achieves a \emph{lower} score on \textsc{ClassicalMechanicsV1} than on
JEEBench empirically demonstrates that our generated problems
successfully capture---and rigorously stress-test---this high tier of
reasoning complexity.

\section{The Complexity Blueprint: \textit{A Priori} Formula Injection}
\label{app:apriori}

A potential concern regarding our \emph{Complexity Blueprint}
(Section~5.3) is whether the linear correlation between formula count
and code length is tautological by construction.  We clarify the
conceptual distinction here.

In standard synthetic generation pipelines, question and solution
generation are decoupled; the number of formulas used can only be
extracted \emph{post-hoc}, after the solution is already generated.
IPG takes the exact opposite route: specific formulas are explicitly
provided as \emph{inputs} during the question generation phase
(Figure~1 and Appendix~C.3, \texttt{sys\_call2}).  This fundamentally
shifts the number of formulas from a passive observation to a tangible,
\emph{a priori} knob for explicitly controlling problem complexity.

The strong linear correlation ($R^{2} \approx 0.953$, Figure~2,
Section~5.3) therefore serves to \emph{empirically} reinforce that our
input-side constraints successfully translate into verifiable
output-side complexity---a non-trivial result, since the LLM could in
principle produce code that invokes fewer or more formulas than
instructed.  The linearity confirms that the agent faithfully respects
the imposed formula budget and does not inject spurious logic.  

\section{Physical Plausibility: Multi-Layered Verification Mechanisms}
\label{app:plausibility}

Code execution in Phase~III verifies numerical correctness and unit
consistency, but does not by itself constitute a full symbolic physics
engine.  We clarify here that IPG incorporates four complementary
mechanisms to enforce physical plausibility.

\begin{enumerate}

  \item \textbf{Fundamental constraint checks (Section~3.4).}  The
  execution-based verifier immediately discards problems whose solutions
  violate basic physical reality---for example, ensuring mass $m > 0$,
  friction coefficients $\mu \in [0,1]$, and time $t > 0$ (referred to
  as ``Physical Sanity'' in Section~3.4).

  \item \textbf{Plausible variable ranges (Section~3.2,
  ``Underlying Principle Extraction'').}  For each variable
  involved in a seed question, Phase~I constructs a Variable
  Dictionary $\mathcal{V} = \{(v_i, u_i, R_i)\}_{i=1}^{n}$ that
  associates every variable with a physically admissible range
  $R_i$.  The LLM is restricted to sampling values exclusively
  within these realistic bounds during Phase~II (Constrained
  Generation).

  \item \textbf{Formula grounding via executable axioms
  (Section~3.1.2).}  Generation is strictly grounded by providing
  relevant physical equations as explicit input prompts through the
  Chapter Dictionary and Available Formula Library mechanisms
  (Figure~1).  The LLM cannot invoke formulas outside this
  domain-validated library.

  \item \textbf{Inherent logic verification (Section~3.4).}  When
  generating solution code in Phase~III, the agent directly applies
  established physical formulas as Python functions.  Because the
  executable code strictly follows these verified mathematical
  representations of physics, it inherently enforces the mathematical
  structure of the physical laws it implements.

\end{enumerate}

This multi-layered approach guarantees a significantly higher rate of
valid, solvable questions compared to a purely zero-shot generation
approach.  We acknowledge, as noted in Section~6 (Limitations), that
residual semantic inconsistencies remain possible for highly complex
scenarios ($N \geq 5$ interacting formulas), and that future work could
integrate formal constraint solvers (e.g., Z3) to enforce higher-order
conservation laws more rigorously.

\section{Domain-Specific Challenges for Math-Focused Augmentation
Methods}
\label{app:domainspecific}

This section provides a theoretical analysis of why methods designed for
abstract mathematical reasoning face inherent challenges when adapted to
physics problem generation---a discussion that contextualises IPG's
design choices relative to MathGenie \cite{mathgenie_acl24},
MetaMath, and Evol-Instruct.

Physics problem-solving fundamentally differs from abstract mathematical
reasoning in requiring the integration of conceptual understanding with
formal mathematics \cite{kuo2013}.  Students blend conceptual and formal
mathematical reasoning when solving physics problems---a cognitive
process distinct from pure symbolic manipulation in mathematics.  This
blending includes grounding equations in physical scenarios,
interpreting intermediate steps through physical principles, and
validating solutions against conservation laws and physical constraints.

The method-specific constraints of existing approaches create inherent
challenges for physics adaptation.

\paragraph{MathGenie's back-translation assumption.}
MathGenie augments solutions and back-translates them into questions,
assuming solution-to-question reversibility.  This works for
mathematics where symbolic manipulation is bidirectional, but breaks
down in physics where multiple distinct physical scenarios can yield
identical mathematical equations---for example, free fall, inclined
plane motion, and projectile motion all reduce to kinematic equations
under specific conditions.

\paragraph{MetaMath's answer-matching verification.}
MetaMath uses rejection sampling that filters solutions based on answer
correctness: diverse reasoning paths are generated and only those with
correct answers are retained.  However, this verification cannot detect
physically invalid solutions that happen to produce numerically correct
answers---for example, solutions that violate energy conservation but
arrive at the correct final velocity through compensating errors.

\paragraph{Evol-Instruct's linguistic evolution.}
Evol-Instruct creates instruction complexity through linguistic prompts
(adding constraints, deepening, concretizing), as demonstrated in
WizardLM.  This approach increases narrative complexity without adding
domain-specific grounding---the evolution is orthogonal to physical
validity.

\paragraph{Evidence from scientific reasoning systems.}
Even SciAgent \cite{li2025sciagent}, explicitly designed for scientific
reasoning, requires entirely separate ``Worker Systems'' for mathematics
versus physics.  The Physics Worker specifically incorporates
``conceptual modeling'' and ``diagram interpretation'' capabilities that
are absent from the Math Worker---capabilities that math-focused
augmentation methods do not address.

\section{Complexity Distribution, Filtering Details, and Corrigendum}
\label{app:complexity}

\subsection{Pre- and Post-Filtering Dataset Sizes}

As stated in Section~5, we initially generated 1,415 candidate problems
and pruned 80 instances ($<6\%$) that required fewer than two deductive
steps, yielding the final corpus of 1,335 problems.  The complexity
distribution in Table~3 reflects this post-filtering dataset.  We note
a typographical error in the submitted version: the ``Total'' cell in
Table~3 incorrectly read 1,335 and has been corrected to 1,415 (the
pre-pruning candidate count) in the revised manuscript.

\subsection{Interpretation of the Complexity Distribution}

While the dataset centres at a mean of 3.05 formulas per problem, IPG
successfully generated 260 problems (19.4\% of the dataset) requiring
4--6 formulas, demonstrating that the framework can produce substantial
multi-step reasoning chains.  The distribution centres at 3 formulas
because this represents the optimal balance for an undergraduate-level
mechanics scope.

Problems requiring 3 formulas represent genuine multi-step reasoning
that integrates concepts across chapters: as shown in Table~4
(Section~5.2), the number of unique formulas utilised per chapter
significantly exceeds its native library size, confirming that
cross-domain mixing is prevalent throughout the corpus.  This level of
complexity substantially exceeds the simple pattern matching often
observed in general Question Generation literature and aligns with the
difficulty of standard physics benchmarks.

The IPG framework's design is deliberately flexible: by adjusting the
formula count constraints in Phase~II (``Difficulty Control,''
Section~3.3), the complexity distribution can be explicitly controlled.
The current distribution reflects a pedagogically appropriate spread for
undergraduate mechanics, but the framework can generate
higher-complexity problems when required.

\subsection{Domain-Specific Concentration of Low-Complexity Instances}

As discussed in Section~5.6, the concentration of zero-formula problems
exclusively in the Centre of Mass chapter and of single-formula problems
in Rigid Body Dynamics is not indicative of a systematic generation
failure.  Rather, it reflects the inherent structure of these domains:
Centre of Mass problems frequently reduce to direct coordinate or
weighted-average calculations, and many Rigid Body Dynamics scenarios
are governed by definitional relations such as $\tau = I\alpha$ and
$L = I\omega$.  When a scenario directly provides inertia and angular
velocity, solving for angular momentum becomes a valid but trivial task.

In contrast, Newton's Laws and Work, Power, and Energy produced very few
pruned instances (5 and 8 problems respectively), as these domains
naturally encourage formula coupling, requiring multiple interacting
quantities such as mass, force, and acceleration.  These chapters
therefore serve as more robust sources for the multi-step reasoning
chains necessary for effective domain adaptation.

The tendency to encounter the uniqueness constraint as a primary failure
mode---particularly in domains where candidate problems fail to be
meaningfully distinct---is reflected in the expansion ratios of
Table~2: Centre of Mass achieves only $6.24\times$ expansion (181
problems from 29 seeds), against targets of $9.31\times$ and
$9.52\times$ for Chapters~5 and~7 respectively.

\section{Model-Agnosticism of the IPG Framework}
\label{app:modelagnostic}

The proof-of-concept reported in this paper exclusively utilises Gemini
models (Gemini~2.5~Flash for generation, Gemini~3 for evaluation).
While this allowed thorough validation of the IPG framework, we
recognise that model diversity would strengthen the generality of our
findings.  The IPG framework is model-agnostic by design: any capable
instruction-following LLM can be substituted in the generation pipeline,
provided it can produce structured JSON outputs and valid Python code.
Multi-model generation and evaluation is noted as important future work.

\end{document}